\documentclass[10pt,twocolumn,letterpaper]{article}

\usepackage{cvpr}              
\definecolor{cvprblue}{rgb}{0.21,0.49,0.74}
\usepackage[pagebackref,breaklinks,colorlinks,allcolors=cvprblue]{hyperref}

\def\paperID{***} 
\def\confName{CVPR}
\def\confYear{2025}

\title{MOGeo: Beyond One-to-One Cross-View Object Geo-localization}

\author{Bo Lv, Qingwang Zhang, Le Wu, Yuanyuan Li, Yingying Zhu\thanks{Corresponding author.}\\
College of Computer Science and Software Engineering, Shenzhen University\\
{\tt\small \{2250271009, zhangqingwang2022, 2300271073, 2200271059\}@email.szu.edu.cn, zhuyy@szu.edu.cn}
}

\begin{document}
\maketitle
\begin{abstract}
 Cross-View Object Geo-Localization (CVOGL) aims to locate an object of interest in a query image within a corresponding satellite image. Existing methods typically assume that the query image contains only a single object, which does not align with the complex, multi-object geo-localization requirements in real-world applications, making them unsuitable for practical scenarios. To bridge the gap between the realistic setting and existing task, we propose a new task, called Cross-View Multi-Object Geo-Localization (CVMOGL). To advance the CVMOGL task, we first construct a benchmark, CMLocation, which includes two datasets: CMLocation-V1 and CMLocation-V2. Furthermore, we propose a novel cross-view multi-object geo-localization method, MOGeo, and benchmark it against existing state-of-the-art methods. Extensive experiments are conducted under various application scenarios to validate the effectiveness of our method. The results demonstrate that cross-view object geo-localization in the more realistic setting remains a challenging problem, encouraging further research in this area. 
\end{abstract}

\section{Introduction}
\label{sec:intro}


Cross-View Geo-Localization (CVGL) aims to determine the geographical location of a query image (e.g., ground or drone image) from geo-tagged reference images (satellite-view images) without relying on GPS or other positioning devices. This technique has broad applications in areas such as autonomous driving ~\cite{hane20173d}, urban navigation~\cite{li2019cross,mirowski2018learning}, smart city management~\cite{hong2021multimodal,sun2019streaming}, and disaster monitoring~\cite{chini2008exploiting,kumar2012environmental}.

\begin{figure}
    \centering
  \includegraphics[width=1\linewidth]{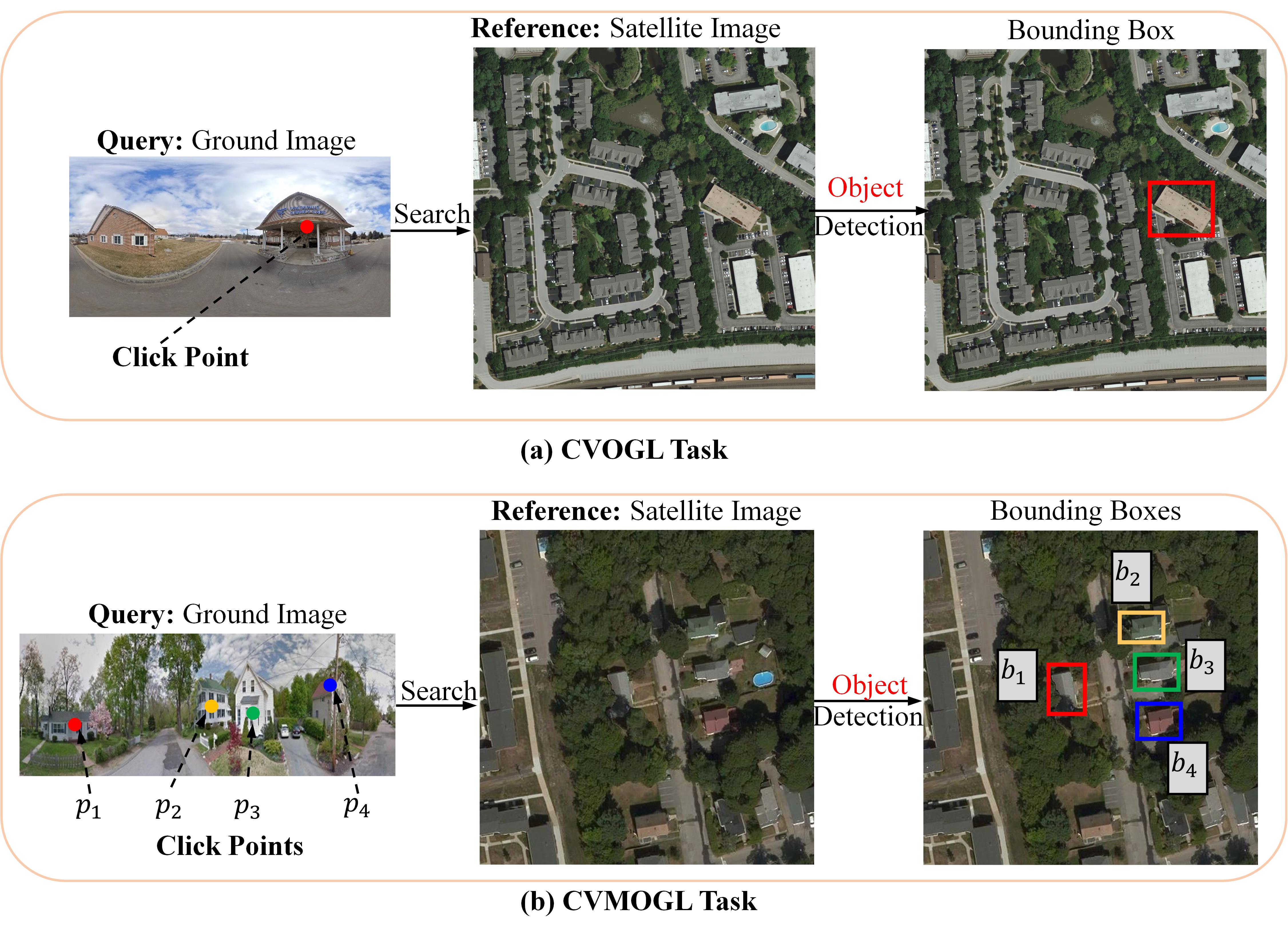}
  \caption{Comparison of cross-view object geo-localization in single-object and multi-object scenarios. Click points represent query objects, while bounding boxes in geo-tagged satellite images indicate location information. Points and bounding boxes of the same color form an object pair, such as $p_1$ and $b_1$, where $b_1$'s geographic location is considered the position of $p_1$.
}
  \label{fig:CVMOGL}
\end{figure}

In recent years, significant research progress has been made in the CVGL task. The research direction has expanded from center alignment of cross-view images to non-center alignment~\cite{zhu2021vigor}, from coarse-grained alignment to fine-grained alignment~\cite{FG2_2025_CVPR}, and from supervised cross-view geo-localization to unsupervised geo-localization methods~\cite{li2024unleashing}. At the same time, research has also shifted from panoramic images to images with limited field of view~\cite{Shugaev_2024_WACV}. The overall trend is gradually moving toward fine-grained localization and closer alignment with real-world scenarios. However, when it comes to localizing objects of interest, such as buildings, within a query image, traditional CVGL methods face significant challenges and limitations, as they are restricted to estimating geo-localization at the image level. To address this problem, the task of Cross-View Object Geo-Localization (CVOGL) is proposed. As illustrated in Fig.~\ref{fig:CVMOGL}(a), the CVOGL task enables geo-localization at the object level. However, it idealizes the application scenario to a single-object setting, focusing solely on the geo-localization of a single object, which contrasts with real-world query images that typically contain multiple objects, such as buildings, roads, and bridges. This limitation undermines its ability to meet the fine-grained geo-localization demands of practical applications, thereby restricting the CVOGL task's effectiveness in real-world scenarios.

 To overcome the limitation of CVOGL, we propose a new task called Cross-View Multi-Object Geo-Localization (\textbf{CVMOGL}). As illustrated in Fig.~\ref{fig:CVMOGL}(b), 
given a query image containing multiple query objects (represented by click points), matching objects (represented by bounding boxes) are located in a geo-tagged satellite image. Points and bounding boxes of the same color in the image represent matched object pairs, and the geographic information of the bounding boxes is considered as the geo-localization of the corresponding query objects. For example, $b_1$ in the satellite image represents the location of the query object $p_1$.
This makes CVMOGL bridge the gap between current research and practical applications, better aligning with practical needs, and enhancing the applicability of CVOGL.

 However, the CVMOGL task presents several new challenges beyond existing settings: (1) CVOGL is a special case of CVMOGL, so existing CVOGL methods are not suitable for the CVMOGL task. CVGL methods provide geo-localization at the image level, failing to offer geo-localization at the object level. (2) In addition to localizing multi-objects in satellite images, it is also necessary to establish the correspondence between each query object (represented by point) and the reference object (represented by a bounding box) to ensure the accuracy of the geo-localization results. 
 
 To advance the research on the CVMOGL task, we first propose a benchmark, called \textbf{CMLocation} (CMLocation-V1 and CMLocation-V2). CMLocation includes a total of 63,888 object instances across 25,520 pairs of query and reference images, providing rich data support for CVMOGL research. Especially, the CMLocation-V2 dataset, with non-center and non-north alignment and varying resolutions, is more representative of real-world scenarios and presents greater challenges.
 
\begin{figure}[t]
  \centering
  \includegraphics[width=0.85\linewidth]{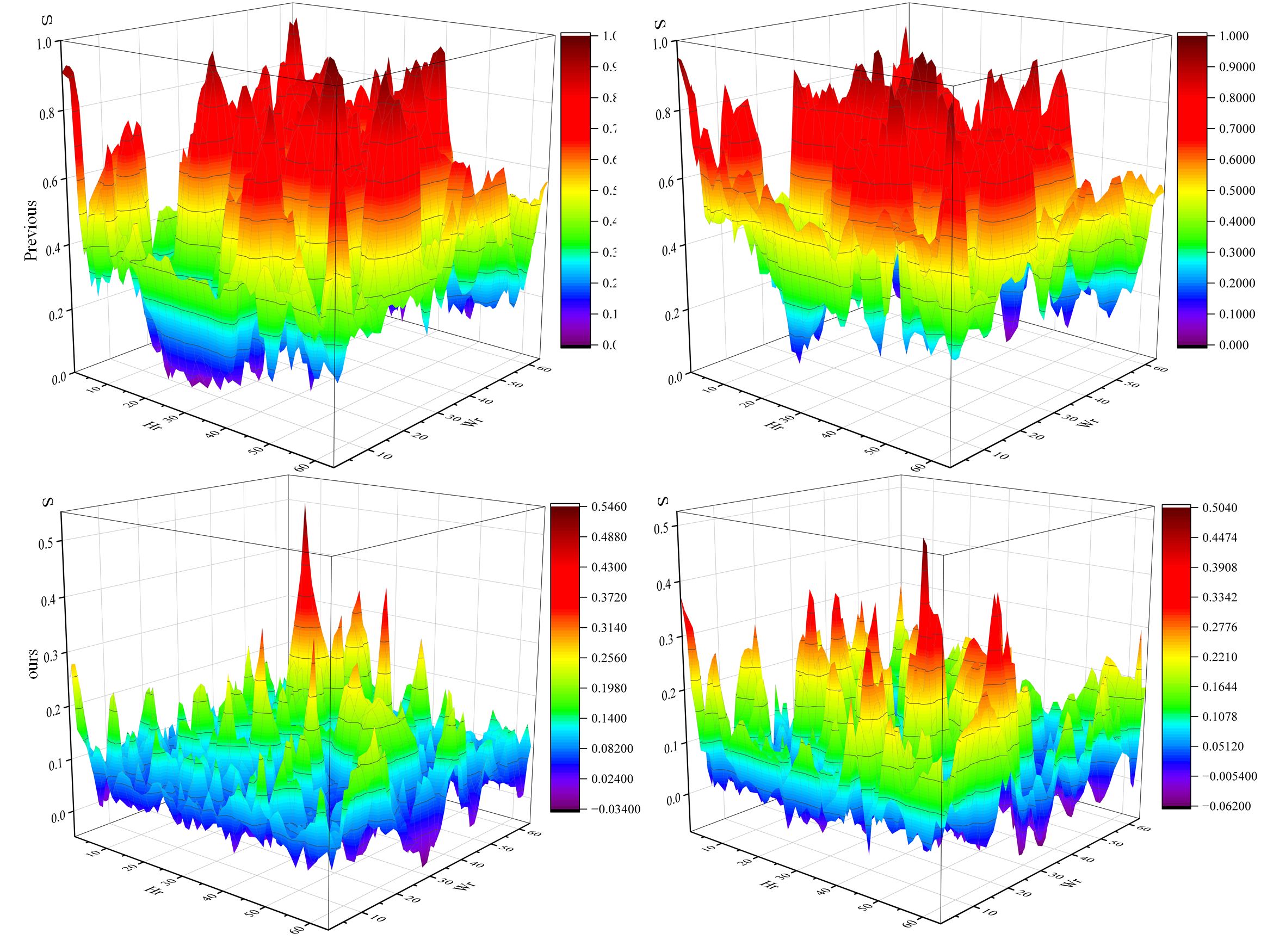}
   \caption{Attention map comparison: previous vs. ours. Previous smooth positional encodings lead to diffuse attention maps, whereas our proposed attention map successfully concentrates on the target location to provide highly discriminative features.
}
   \label{fig:different_attention2}
\end{figure}

Moreover, to further advance the research on the CVMOGL task, we propose an end-to-end cross-view multi-object geo-localization method, \textbf{MOGeo}. MOGeo includes a dual-branch feature encoding architecture, Multi-Object Position Encoding (MOPE), cross-view multi-feature fusion (CVMF), and training objective function. The dual-branch encoder independently extracts visual features from each view image. As shown in Fig.~\ref{fig:different_attention2}, previous smooth positional encodings (e.g., Gaussian or Euclidean) create diffuse attention maps that hinder precise focus on the query object. To resolve this spatial ambiguity, we propose a sharp, non-smooth delta-like impulse encoding inspired by the Dirac delta function~\cite{lighthill1958introduction}. This method provides a highly discriminative positional prior for each query point. We realize this idea with our novel MOPE module, ultimately boosting the model's localization accuracy. Additionally, to fuse cross-view multi-object features, we design a cross-view multi-feature fusion module that emphasizes the importance of attention maps. Furthermore, based on prior knowledge, objects in different images, or different objects within the same image exhibit distinct attention distribution features. To enhance the distinction in attention distribution features, we design a novel optimization function. Finally, the fused features are processed through multiple detection heads to predict the corresponding locations of the query objects within the reference image.

Our main contributions can be summarized as follows:
\begin{itemize}
    \item To break the limitation of CVOGL, we propose a new task of Cross-View Multi-Object Geo-Localization (CVMOGL). CVMOGL addresses the multi-object geo-localization problem, offering a more challenging and realistic CVOGL setting.
    \item To bridge the gap between the realistic setting and existing datasets, we first construct a benchmark, called CMLocation( CMLocation-V1 and CMLocation-V2). This benchmark includes a total of 63,888 object instances in 25,520 pairs of query and reference images, providing rich data support for CVMOGL research.
    \item To promote the research on the CVMOGL task, we propose a cross-view multi-object geo-localization method, called MOGeo. We evaluate
 MOGeo and other methods on the CMLocation benchmark. The evaluation results show that MOGeo achieves state-of-the-art performance, fully demonstrating its effectiveness. 
\end{itemize}

\section{Related work}
\subsection{Cross-view Image Geo-localization}
Cross-view image geo-localization has made significant progress in both methods~\cite{chen2025multi, lv2024direction, chen2025cross, li2025vageo, shore2025spagbol} and datasets~\cite{ye2024cross, workman2015wide, sun2023cross, chen2025cross, huang2024cv, xia2024cross, zhai2017predicting}.

In terms of datasets, the CVUSA~\cite{workman2015wide, zhai2017predicting} and CVACT~\cite{liu2019lending} datasets use ground-view images as queries matched with high-resolution satellite imagery, while University-1652~\cite{zhenguniversity1652} and SUES-200~\cite{zhu2023sues} datasets support drone geo-localization. However, the aforementioned datasets mainly consider one-to-one retrieval, which may not align well with real-world scenarios. To address this, VIGOR~\cite{zhu2021vigor} allows non-center-aligned image pairs for more realistic retrieval. However, these datasets provide geo-localization at the image level, but do not offer localization at the object level.

In terms of localization methods, Shi et al.~\cite{shi2019spatial} applied polar transformations to align satellite-view and ground images, while Regmi et al.~\cite{regmi2019bridging} used conditional GANs to synthesize ground-level images from satellite-view inputs. Lu et al.~\cite{lu2020geometry} and Toker et al.~\cite{toker2021coming} incorporated geometric priors and multi-task learning to bridge the cross-view domain gap. More recently, methods based on Swin Transformers~\cite{lv2024direction, shi2022accurate} and unsupervised learning~\cite{li2024unleashing} have achieved state-of-the-art results on UAV-view datasets. Chen et al.~\cite{chen2025cross} introduced GeoSSK, which improves geo-localization accuracy between ground and satellite-view images using cross-view semantic similarity learning and knowledge distillation. Zhang et al.~\cite{zhang2024geodtr+} proposed GeoDTR+, which enhances cross-region generalization by modeling geometric layouts and generating hard training samples. However, these studies are limited to image-level localization, restricting their real-world applicability.

\subsection{Cross-View Object Geo-localization}
Cross-view object geo-localization research is still in its early stages, and related studies are relatively limited. The DetGeo method~\cite{sun2023cross} pioneers object-level cross-view geo-localization, and the VAGeo method~\cite{li2025vageo} has further advanced research on CVOGL. More recent works, such as TROGeo~\cite{zhang2025TROGeo} and GeoFormer~\cite{zhang2025Geoformer}, explore this problem from two key aspects: supervision paradigms and object discriminability. However, these methods are primarily restricted to single-object settings, limiting their applicability in real-world scenarios. To bridge the gap between current research and practical applications, we first propose the CVMOGL task. 

\section{The Proposed Benchmark}
To advance research on the cross-view multi-object geo-localization (CVMOGL) task, we construct a cross-view multi-object geo-localization benchmark, named CMLocation. Below, we describe the CMLocation benchmark in detail.

\begin{figure}[t]
  \centering
  \includegraphics[width=1\linewidth]{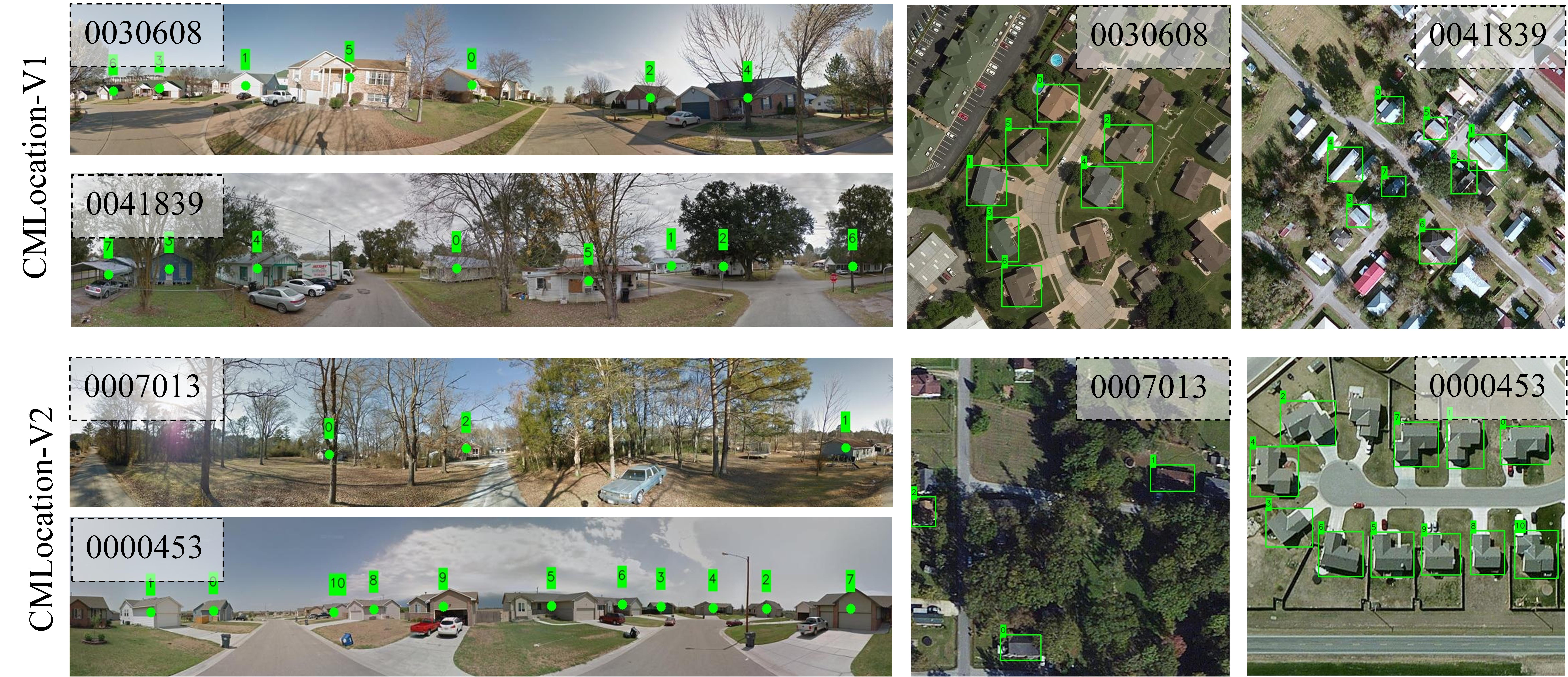}
   \caption{Examples from the  CMLocation-V1 and  CMLocation-V2 datasets. 
 CMLocation-V1 is curated under strict alignment principles—specifically center alignment and northward orientation—which ensure a consistent spatial distribution across all instances. 
By contrast,  CMLocation-V2 does not impose these constraints, resulting in more heterogeneous spatial configurations.}
   \label{fig:CVMOGL_V1_V2_sample}
\end{figure}

\noindent \textbf{Dataset Collection and Object Annotation.} The emphasized properties of the cross-view multi-object geo-localization dataset are two major elements: cross-view and multi-object. Therefore, based on the cross-view image datasets CVUSA, we manually annotate visible objects in both query and reference images using the LabelImg tool. We represent the annotated bounding boxes in the query image with a random point inside the box (as a click point). The corresponding relationships between query points in the query image and objects in the reference image are manually determined and recorded as object correspondence labels. As shown in Fig.~\ref{fig:CVMOGL_V1_V2_sample}, points and boxes with the same label number form a matching relationship. Finally, we obtain the CMLocation-V1 dataset. 
However, because the query and reference images in CMLocation-V1 are arranged in a north-aligned and center-aligned fashion (which facilitates determining object correspondences during manual annotation), this setup differs significantly from real-world application scenarios. Therefore, we applied random cropping, flipping, and scaling to the reference images in  CMLocation-V1 so that the query and reference images would no longer share the same center point or fixed orientation. This process resulted in the CMLocation-V2 dataset, which is more realistic and challenging for real-world applications. More details about the dataset can be found in the appendix.

\begin{table}[ht]
\caption{Size and partitioning information of the  CMLocation-V1 and  CMLocation-V2 datasets.
}
\label{tab:dataset_split}
\small
\resizebox{\columnwidth}{!}{%
\begin{tabular}{ccccc}
\toprule\toprule
Data name & Split & Query & Reference & Instances \\
\midrule
\multirow{3}{*}{ CMLocation-V1} & total & 12760 & 12760 & 31944 \\
 & training & 8371 & 8371 & 20931 \\
 & validation & 2093 & 2093 & 5209 \\
 & test & 2296 & 2296 & 5804 \\            
\midrule
\multirow{3}{*}{ CMLocation-V2} & total & 12760 & 12760 & 31944 \\
 & training & 8371 & 8371 & 20931 \\
 & validation & 2093 & 2093 & 5209 \\
 & test & 2296 & 2296 & 5804 \\
\bottomrule\bottomrule
\end{tabular}
}
\end{table}

\noindent \textbf{Dataset Split.} Table~\ref{tab:dataset_split} summarizes the data scale and splits of CMLocation-V1 and CMLocation-V2. Both versions contain 12{,}760 query-reference image pairs and share the same set of query objects (31{,}944 in total), providing rich data support for CVMOGL research.

\begin{figure}[t]
  \centering
  \includegraphics[width=1\linewidth]{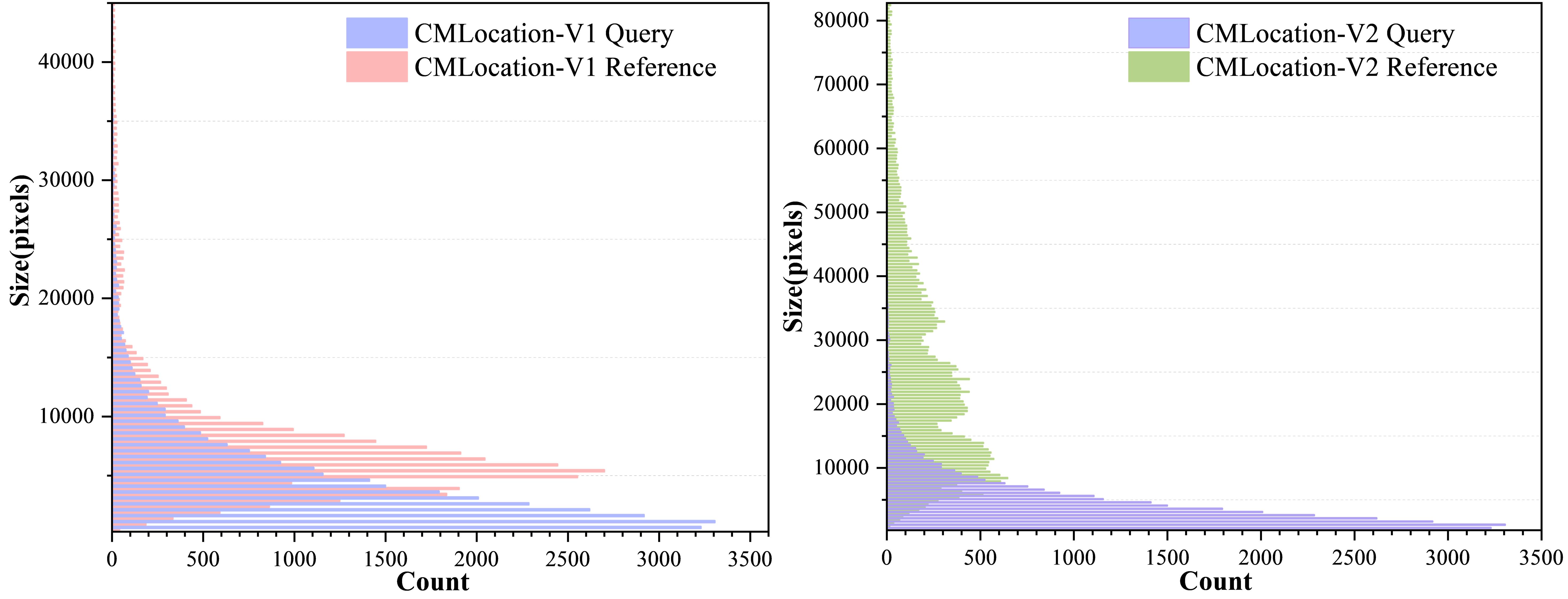}
   \caption{The size distributions of objects in the  CMLocation-V1 and  CMLocation-V2 datasets.}
   \label{fig:CMLocation_distribution}
   \vspace{-8mm}
\end{figure}

To analyze object size distributions under different views, we compute the bounding box sizes of query and reference objects in both versions, as shown in Fig.~\ref{fig:CMLocation_distribution}. The results reveal a wide range of object sizes in both datasets, with query-side objects generally appearing smaller than their counterparts in the reference images. In CMLocation-V2, the scale discrepancy becomes even more pronounced due to the more extreme viewpoint variations, presenting greater challenges for robust model training.

\noindent \textbf{Evaluation Settings.} The \(\text{acc@}t\) metric in \cite{sun2023cross} evaluates single-object localization but fails to assess overall multi-object localization accuracy in the CVMOGL task. Therefore, we introduce the concept of image localization accuracy.
The calculation formula is:
\begin{equation}
\text{accI@}t = \frac{1}n \sum_{i=1}^{n} \psi_i(t)
\end{equation}
\begin{equation}
\psi_i(t) = 
\begin{cases} 
0, & \text{if } \exists j \in \{1, 2, \dots, m_i\} \text{, IoU}(b_j, b_j^*) > t \\
1, & \text{else}
\end{cases}
\end{equation}
where \( n \) is the number of query images, \( m_i \) is the number of query objects in the \(i\)-th image, \( b_i \) is the predicted bounding box, and \( b_i^* \) is the ground truth bounding box. In these metrics, \(\text{accI@}t\) denotes the proportion of images in the dataset where the objects are correctly localized. An image is deemed correctly localized when the Intersection over Union (IoU) between all predicted and ground truth bounding boxes exceeds \(t\). To ensure a fair comparison, we adopt the same experimental setup as in \cite{sun2023cross}, setting the threshold \(t\) to 0.25 and 0.5, respectively.

\label{sec:4_methodology}
\begin{figure*}[t]
  \centering
  \includegraphics[width=0.9\linewidth]{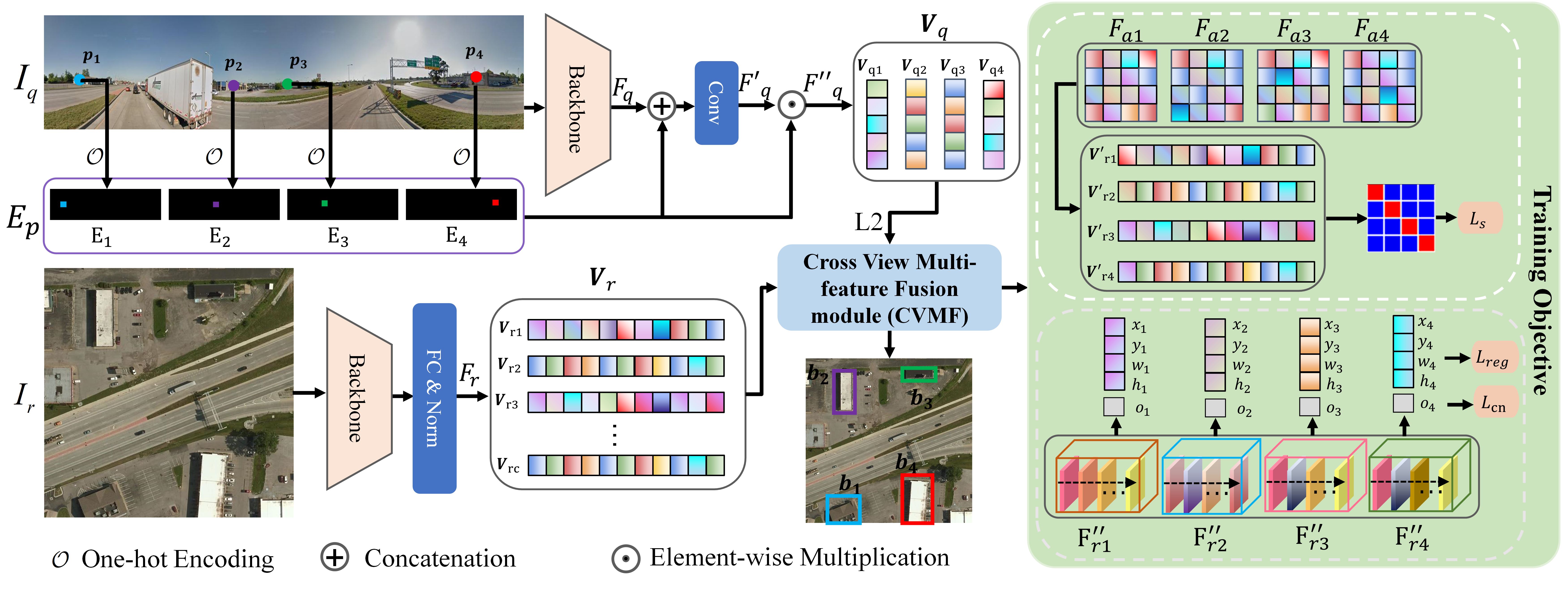}
   \caption{ Overview of our proposed MOGeo model. Our method takes as input a query image that contains an arbitrary number of objects of interest (either ground-view images or drone-view images) and a reference image that contains the objects to be localized (satellite image). The input images are processed by our position encoding module, MOPE, to extract feature information for each query object. Then, query object features are fused with the reference image to obtain fused features \( \mathbf{F''}_r \), which are fed into the detection head to produce localization results. In the figure, query points and bounding boxes with the same color represent object pairs.
} 

   \label{fig:MOGeo_model}
\end{figure*}

\begin{figure}[t]
  \centering
  \includegraphics[width=1\linewidth]{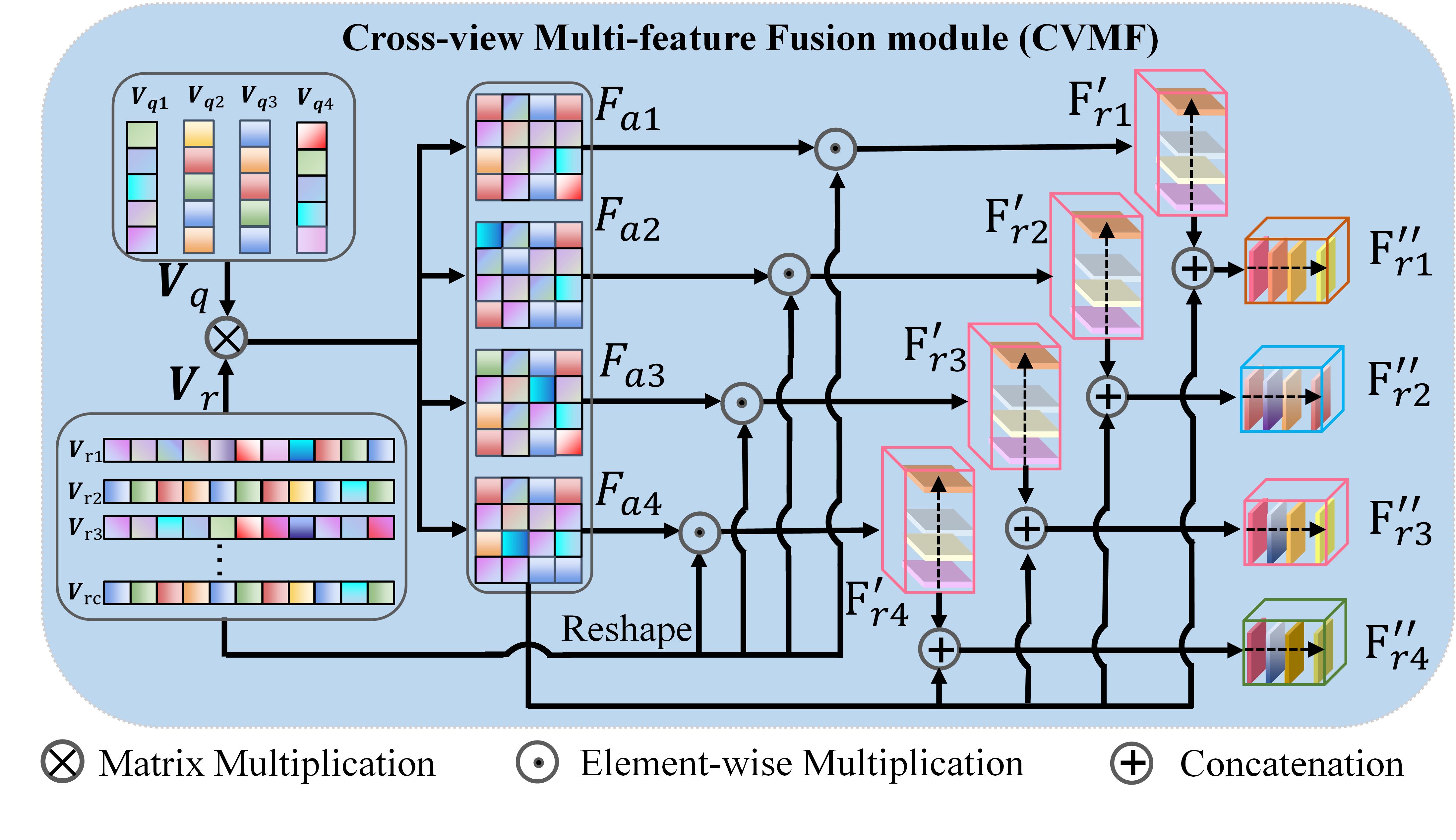}
   \caption{Cross-view multi-object feature fusion module. As shown in the figure, this module sequentially fuses query object features with the reference image to obtain an attention map for each object, and finally combines fused features with the attention map to further enhance localization in the reference image.
}
   \label{fig:CVMOF_and_Loss}
\end{figure}

\section{ The Proposed Method}
In this section, we aim to locate objects in cross-view images, typically represented by bounding boxes. To achieve this, we propose a novel detection-based CVMOGL method, called MOGeo, as shown in Fig. \ref{fig:MOGeo_model}. MOGeo consists of four main components: a dual-branch image feature extraction network, multi-head query position encoding, a multi-feature fusion module, and an optimization function. In the following, we will provide a detailed explanation of each component.

 \subsection{Problem Statement}

A set of query points and reference objects is represented as \( X = \{(q_i, r_i, p_j, b_j)_{j=1}^{m_i}\}_{i=1}^{n} \), where the query image \( q_i \) and reference image \( r_i \) form a matching image pair, and the query point \( p_j \) and bounding box \( b_j \) form a matching object pair. \( m_i \) represents the number of object pairs contained in the \( i \)-th image pair. Given a pair of query and reference images, and the positions of the query points in the query image, our goal is to identify the corresponding bounding boxes in the reference image. Thus, the problem is formulated as \( (q_i, r_i, p_j)_{j=1}^{m_i} \rightarrow b_j \).

For convenience, we choose the \(i\)-th sample from dataset \( X \) as the input data: \( x_i = (q_i, r_i, p_j, b_j)_{j=1}^{m_i}, \, q_i \in \mathbb{R}^{H_q \times W_q \times 3}, \, r_i \in \mathbb{R}^{H_r \times W_r \times 3} \). We denote the \(i\)-th query image and reference image as \( I_q \) and \( I_r \), respectively. \( P = \{p_1, p_2, \dots, p_{m_i}\} \) represents the query points, each corresponding to a ground truth (GT) bounding box, denoted as \( \{b_1, b_2, \dots, b_{m_i}\} \), where \( b_j = (x^*_{j}, y^*_{j}, w^*_{j}, h^*_{j}) \).

\textbf{Query and Reference Image Representation.}
To ensure fairness in subsequent experimental comparisons, we choose the same backbone architecture for feature extraction as in the works of \cite{sun2023cross, li2025vageo}. For the query image \( I_q \), we use ResNet18 to extract features, selecting the downsampled \( 16 \times \) result as the feature representation \( F_q \in \mathbb{R}^{C \times H_q' \times W_q'} \).
For the reference image \( I_r \), we use Darknet53 for feature extraction and select the downsampled \( 16\times \) result. A fully connected operation is applied to obtain a more expressive feature \( F_r \). We reshape \( F_r \) to \( \mathbf{V}_r = \{ v_{r1}, v_{r2}, \dots, v_{rc} \} \), where \( \mathbf{V}_r \in \mathbb{R}^{4096 \times 512} \), for subsequent feature fusion.

\subsection{ Multi-Head Query Object Encoding}
To support multi-object geo-localization in query images, we propose a multi-head query object position encoding (\textbf{MOPE}) module based on feature post-processing. Inspired by the Dirac delta function, we implement our impulse-based positional encoding as a one-hot mask, ensuring a sharp and discriminative spatial prior. Specifically, after obtaining the feature map \( F_q \) of the query image, position encoding is performed by creating \( m \) binary masks \( \{E_1, E_2, \dots, E_m\}_{j=1}^m \subset \{0,1\}^{H'_q \times W'_q} \), each encoding the position of a query point using the one-hot principle. The mathematical representation is as follows:
\begin{equation}
E_j(h,w) =
\begin{cases}
1, & \text{if } (h, w) = \lfloor \varphi(p_j) \rfloor \\
0, & \text{otherwise}
\end{cases}
\end{equation}
\noindent where \( \varphi: \mathbb{R}^2 \to \mathbb{R}^2 \) denotes the transformation that maps the query point coordinates to the feature map coordinates, and \( \lfloor \cdot \rfloor \) represents the floor operation. After obtaining the absolute position information of the query objects, we fuse each position encoding \( E_j \) with the visual feature \( F_q \) via channel concatenation, followed by a \( 1 \times 1 \) convolution to transform the combined feature, resulting in the enhanced feature \( F_{qj}' \in \mathbb{R}^{C \times H'_q \times W'_q} \):
\begin{equation}
\ F_{qj}'  = \text{Conv}_{1 \times 1} \left( \left[ F_q \| E_j \right] \right).
\end{equation}
\noindent Here, \( \text{Conv}_{1\times1} \) denotes a convolution with a \( 1 \times 1 \) kernel, and \( [\cdot \| \cdot] \) represents concatenation along the channel dimension. For simplicity, we denote \( \{ F_{q1}', F_{q2}', \dots, F_{qm}' \} \) as \( F_q' \). Through the above operations, we obtain the query image information with the initial fusion feature \( F_q' \). However, a significant channel imbalance exists between \( F_q \) and \( E_j \), which may lead to the position information being overshadowed or weakened by the visual feature information. To mitigate the potential dilution of position information by semantic features, we propose a position-driven feature enhancement strategy. By expanding the position mask along the channel dimension and performing element-wise multiplication, we obtain the position-enhanced fused feature \( F_q'' \):
\begin{equation}
\{ F_{q1}'', F_{q2}'', F_{q3}'', \dots, F_{qm}'' \} = F_q' \odot \{ {E}_{1}, {E}_{2}, \dots, {E}_{m} \}.
\end{equation}
Here, \( \odot \) denotes element-wise multiplication. The fused feature \( \{ F_{q1}'', F_{q2}'', F_{q3}'', \dots, F_{qm}'' \} \) is then transformed and pooled to produce \( m \) query feature vectors, denoted as \( \mathbf{V}_q = \{ v_{q1}, v_{q2}, \dots, v_{qm} \} \), where \( \mathbf{V}_q \in \mathbb{R}^{m \times d} \), with \( d = 512 \).

\begin{table*}
\centering
\small
\caption{Performance (\text{accuracy}\%) comparison of different approaches on our dataset CMLocation.}
\label{tab:ours_method_CVMOGL_preformace}
\begin{tabular}{lccccccccc}
\hline\hline
\multirow{2}{*}{Method} & \multicolumn{4}{c}{Test} & \multicolumn{4}{c}{Validation} \\
\cmidrule(lr){2-5} \cmidrule(lr){6-9}
  & acc@0.25 & acc@0.5 & accI@0.25 & accI@0.5 & acc@0.25 & acc@0.5 & accI@0.25 & accI@0.5\\
\hline
 CMLocation-V1 \\
\hline
FRGeo\cite{zhang2024FRgeo} & 8.06 & 1.05 & 2.74 & 0.17 & 7.20 & 0.92 & 2.53 & 0.05 \\
GeoDTR+\cite{zhang2024geodtr+} & 7.44 & 0.45 & 3.83 & 0.17 & 8.62 & 0.77 & 4.30 & 0.53 \\
Sample4Geo\cite{deuser2023sample4geo} & 29.19 & 2.31 & 16.46 & 1.44 & 30.68 & 2.36 & 19.10 & 1.81 \\
DetGeo\cite{sun2023cross} & 56.03 & 55.00 & 39.68 & 38.32 & 56.29 & 55.33 & 39.80 & 38.93 \\   
VAGeo\cite{li2025vageo} & 57.79 & 57.06 & 43.03 & 42.33 & 57.04 & 56.50 & 42.66 & 42.09 \\
\textbf{MOGeo} & \textbf{63.85} & \textbf{62.92} & \textbf{46.66} & \textbf{45.60} & \textbf{63.52} & \textbf{62.58} & \textbf{46.82} & \textbf{45.44} \\
\hline
 CMLocation-V2\\
\hline
FRGeo\cite{zhang2024FRgeo} & 9.87 & 1.43 & 4.05 & 0.13 & 9.83 & 1.27 & 2.81 & 0.05 \\
GeoDTR+\cite{zhang2024geodtr+} & 11.10 & 1.72 & 3.22 & 0.13 & 11.69 & 1.56 & 3.20 & 0.19 \\
Sample4Geo\cite{deuser2023sample4geo} & 32.43 & 4.86 & 18.78 & 1.22 & 31.37 & 4.15 & 19.10 & 0.86 \\
DetGeo\cite{sun2023cross} & 32.39 & 31.79 & 28.05 & 27.00 & 33.10 & 32.56  & 30.05 & 29.14  \\
VAGeo\cite{li2025vageo} & 33.51 & 33.06 & 29.18 & 28.17 & 34.38 & 33.98  & 31.25 & 30.53  \\
\textbf{MOGeo} & \textbf{37.87} & \textbf{36.63} & \textbf{30.23} & \textbf{28.26} & \textbf{39.16} & \textbf{37.40} & \textbf{34.20} & \textbf{31.58} \\
\hline\hline
\end{tabular}
\end{table*}

\subsection{ CVMF Module}

After obtaining the query object representation vector \( \mathbf{V}_q \) and reference image feature vector \( \mathbf{V}_r \), we fuse the features from both views for object localization in the satellite image. To ensure robust feature matching, both \( \mathbf{V}_q \) and \( \mathbf{V}_r \) undergo \( L_2 \) normalization. Subsequently, matrix multiplication is applied to generate a cross-view attention map, as defined by:
\begin{equation}
\{ \mathbf{V}_p \} = \{ v_{p1}, v_{p2}, v_{p3}, \dots, v_{pm} \} = \{ \mathbf{V}_q \} \times \{ \mathbf{V}_r \}.
\end{equation}
Reshape \( \mathbf{V}_{pj} \) to a tensor of dimensions \( H_r \times W_r \) to obtain the cross-view attention map \( F_{aj} \). The attention map \( F_a = \{ F_{a1}, F_{a2}, \dots, F_{am} \} \) captures the potential corresponding regions of the query objects in the reference image. Each attention feature map \( F_{aj} \) is then weighted and integrated with the satellite image, as defined by the following formula:
\begin{equation}
\{ \mathbf{F'}_r \} = \{ \mathbf{F}_a \} \odot \{ \mathbf{F}_r \}.
\end{equation}
To further enhance the impact of each attention feature \( F_{ai}\) in the fused features, we concatenate the attention features \( F_{ai}\) of each query object with the corresponding fused features \( \mathbf{F'}_r \). Finally, the concatenated features \( \mathbf{F''}_r \) are processed through the prediction heads to generate the bounding box, with the box of highest confidence selected as the final prediction.
 
\subsection{Training Objective }
We design a loss function to optimize MOGeo based on~\cite{sun2023cross}. The loss is defined as:
\begin{equation}
\mathcal{L} = L_{\text{cn}} + L_{\text{reg}} + L_s ,
\end{equation}
where \( L_s \) is the similarity loss between different objects, \( L_{\text{reg}} \) is the regression loss, and \( L_{\text{cn}} \) is the confidence loss. The regression loss minimizes the distance between the predicted bounding box and the GT, while the confidence loss estimates the probability of an object being present in a specific grid cell. The similarity loss computes the differences between attention feature vectors of different query objects. As illustrated in Fig.~\ref{fig:different_attention2}, previous methods suffer from the problem of overly diffuse attention maps. Aiming to further enhance the discriminative power of the attention feature vectors, we utilize this loss to explicitly maximize the distance between non-corresponding query objects, both within a single image and across different images. The calculation formula for \( L_s \) is as follows:
\begin{equation}
L_{\text{s}} = \sum_{i=1}^n\sum_{k=1}^{m_i}\log \left( 1 +  e^{   (d_{\text{pos}} - d_{\text{neg}})} \right) ,
\end{equation}
where \( d_{\text{pos}} \) represents the Euclidean distance between the attention map of the current query object and itself, while \( d_{\text{neg}} \) represents the Euclidean distance between the attention map of the current query object and the attention maps of other query objects. Notably, since the positive distance \( d_{\text{pos}} \) (computed between features of the same object) is already minimized by design, our loss function solely constrains the maximization of distances between different objects.

\section{ Experiment}
\subsection{Experimental Settings}
 Our model is implemented using PyTorch, and the training process is conducted on an NVIDIA V100 GPU. We use the Adam optimizer with an initial learning rate of \( 1 \times 10^{-4} \). The batch size is set to 8, and the model is trained for a total of 24 epochs. We evaluate our method on the  CMLocation dataset (CMLocation-V1,  CMLocation-V2) and CVOGL(CVOGL-SVI,CVOGL-Drone) dataset.
Specifically, CMLocation and CVOGL-SVI are used for the 
\emph{Ground}\textrightarrow\emph{Satellite} scenario, 
while CVOGL-Drone is dedicated to the 
\emph{Drone}\textrightarrow\emph{Satellite} scenario. For more experimental results and analysis, please refer to the Appendix.

\begin{table}[ht]
\centering
\small
\setlength{\tabcolsep}{3pt}
\caption{Performance (\text{accuracy}\%) comparison of different approaches on CVOGL dataset.}
\label{our_method_comparison_on_CVOGL}
\begin{tabular}{llcccc}
\toprule
 \multirow{2}{*}{Method}& \multicolumn{2}{c}{Test} & \multicolumn{2}{c}{Validation} \\
\cmidrule(lr){2-3} \cmidrule(lr){4-5}
 & acc@0.25 & acc@0.5 & acc@0.25 & acc@0.5 \\
\bottomrule
CVOGL-SVI \\
\bottomrule
CVM-Net \cite{hu2018cvm}         &4.73 & 0.51   &5.09 &0.87  \\
TransGeo\cite{zhu2022transgeo}          & 21.17 & 2.88  & 21.67 & 3.25  \\
FRGeo \cite{zhang2024FRgeo}         & 8.12 &1.31  & 7.80 & 0.87  \\
GeoDTR \cite{zhang2023cross}        & 32.37 & 6.06  & 31.53 & 5.31 \\
Sample4Geo\cite{deuser2023sample4geo}    & 8.84 & 1.23   & 11.48 & 1.41 \\
DetGeo \cite{sun2023cross}  & 45.43 & 42.24  & 46.70 & 43.99  \\
VAGeo \cite{li2025vageo}   & 48.21 & 45.22  & 47.56 & 44.42  \\
\textbf{Ours} & \textbf{50.98} & \textbf{47.70} & \textbf{49.51} & \textbf{44.42} \\
\hline
CVOGL-Drone \\
\hline
CVM-Net\cite{hu2018cvm}            & 20.14 & 3.29  & 20.04 & 3.47 \\
TransGeo\cite{zhu2022transgeo}         & 35.05 & 6.37 & 34.78 & 5.42 \\
FRGeo \cite{zhang2024FRgeo}          & 11.41 & 2.67 & 13.22 &2.06 \\
GeoDTR \cite{zhang2023cross}      & 32.48 & 5.65 & 31.13 & 5.53 \\
Sample4Geo\cite{deuser2023sample4geo}  & 7.71& 0.92& 11.16& 1.41 \\
DetGeo\cite{sun2023cross}  & 61.97 & 57.66 & 59.81 & 55.15 \\
VAGeo  \cite{li2025vageo}  & 66.19 & \textbf{61.87}  & 64.25 & \textbf{59.59}  \\
\textbf{Ours} & \textbf{66.39} & 59.71 & \textbf{65.11} & 58.72 \\
\bottomrule
\end{tabular}
\end{table}

\subsection{Comparison with State-of-the-Art Methods}
To ensure a fair comparison with existing retrieval-based localization methods, we divide the satellite reference image into \(128 \times 128\) pixel patches. Each query object is encoded into a feature representation, and the top five most similar patches are retrieved from the reference image. An object is considered successfully localized if at least one retrieved patch has an Intersection over Union (IoU) with the ground truth bounding box exceeding a specified threshold. Otherwise, it is deemed a failure.

Additionally, to compare fairly with existing object detection methods, we convert our dataset into a single-object geo-localization format for evaluation. 

\textbf{(1) Performance of Cross-View Multi-Object Geo-Localization.} 
Table~\ref{tab:ours_method_CVMOGL_preformace} presents a comprehensive comparison of MOGeo with other methods on our CMLocation dataset. MOGeo demonstrates superior performance across both CMLocation-V1 and CMLocation-V2 datasets, particularly in acc@0.25 and accI@0.25 metrics. For example, on the CMLocation-V1 test set, MOGeo surpasses the second-best method, VAGeo, by 6.06\% in acc@0.25, and by 3.63\% in accI@0.25. On the more challenging CMLocation-V2, although overall performance declines compared to CMLocation-V1, MOGeo consistently outperforms strong baselines. It achieves relative gains of 4.78\% and 4.15\% over VAGeo and DetGeo in acc@0.25 and accI@0.25, respectively, on the validation set. These results demonstrate the effectiveness and robustness of MOGeo across varying levels of task difficulty, consistently achieving state-of-the-art (SOTA) performance. The substantial improvement in accI@0.25 and accI@0.50 highlights that MOGeo provides fine-grained object geo-localization, rather than merely estimating rough positions. These results also suggest that existing cross-view image geo-localization methods may not be well-suited for the CVMOGL task.

\textbf{(2) Performance of Cross-View Single-Object Geo-Localization.}  
\noindent The CVOGL task is a special case of the CVMOGL task. To further evaluate the effectiveness of our method, we conducted experiments on the CVOGL dataset. 
The experimental results of different methods are shown in Table~\ref{our_method_comparison_on_CVOGL}. As seen from the table, our method achieves SOTA or competitive performance in the \emph{Ground}\textrightarrow\emph{Satellite} and \emph{Drone}\textrightarrow\emph{Satellite} tasks.
Specifically, in the \emph{Ground}\textrightarrow\emph{Satellite} task, MOGeo achieves 50.98\% in acc@0.25, a 2.77\% improvement over VAGeo and a 5.55\% improvement over DetGeo. 
Similarly, in the \emph{Drone} $\rightarrow$ \emph{Satellite} task, MOGeo achieves SOTA on both the test and validation sets for acc@0.25, while remaining competitive with the best-performing methods on acc@0.50. These results further confirm MOGeo's superior performance in multi-object geo-localization, while maintaining competitive accuracy in the single-object geo-localization setting.

\begin{table}[t]
\centering
\caption{Model performance (\text{accuracy}\%) under different numbers of query objects on the validation set.}
\label{tab:multi_obj_classes_N_performance_validation}
\small
\setlength{\tabcolsep}{0.3pt} 
\begin{tabular}{lcccccc} 
\toprule
\multirow{3}{*}{Method} & \multicolumn{6}{c}{Validation} \\
\cmidrule(lr){2-7} 
 & \multicolumn{2}{c|}{\textbf{I ($N \leq 3$)}} & \multicolumn{2}{c|}{\textbf{II ($3 < N \leq 6$)}} & \multicolumn{2}{c}{\textbf{III ($N > 6$)}} \\
\cmidrule(lr){2-3} \cmidrule(lr){4-5} \cmidrule(lr){6-7}
 & acc@0.25 & acc@0.5 & acc@0.25 & acc@0.5 & acc@0.25 & acc@0.5 \\
\midrule
\multicolumn{7}{c}{CMLocation-V1} \\ 
\midrule
DetGeo  & 61.03 & 59.90 & 49.03 & 48.23 & 40.58 & 40.34 \\
VAGeo   & 62.97 & 62.30 & 47.74 & 47.54 & 42.27 & 42.27 \\
\textbf{Ours}  & \textbf{67.22} & \textbf{66.07} & \textbf{56.74} & \textbf{56.34} & \textbf{53.14} & \textbf{52.90} \\
\midrule
\multicolumn{7}{c}{CMLocation-V2} \\ 
\midrule
DetGeo  & 40.47 & 39.75 & 17.90 & 17.80 & 11.59 & 11.59 \\
VAGeo   & 42.09 & 41.58 & 19.09 & 18.95 & 10.39 & 10.39 \\
\textbf{Ours}  & \textbf{46.37} & \textbf{44.05} & \textbf{25.86} & \textbf{25.31} & \textbf{18.84} & \textbf{18.36} \\
\bottomrule 
\end{tabular}
\end{table}

\textbf{(3) Impact of the Query Object Count on Model Performance.}  
To investigate the effect of query object count on model performance, we divided the CMLocation dataset into three categories based on the number of objects (N) in the query image: $I \ (N \leq 3)$, $II \ (3 < N \leq 6)$, and $III \ (N > 6)$. The results in Table~\ref{tab:multi_obj_classes_N_performance_validation} show that while all methods experience a performance decline with more query objects, our proposed MOGeo consistently achieves state-of-the-art performance in all settings. Notably, MOGeo exhibits superior robustness, and its performance degradation is less severe than that of the baseline methods as N increases. This leads to a widening performance gap in more challenging scenarios. For example, on CMLocation-V2 (acc@0.25), MOGeo’s lead over the second-best method grows from 4.28\% in the simplest category (I) to 7.25\% in the most complex one (III), demonstrating its enhanced capability in handling dense multi-object localization tasks.

\begin{figure}
    \centering
  \includegraphics[width=1\linewidth]{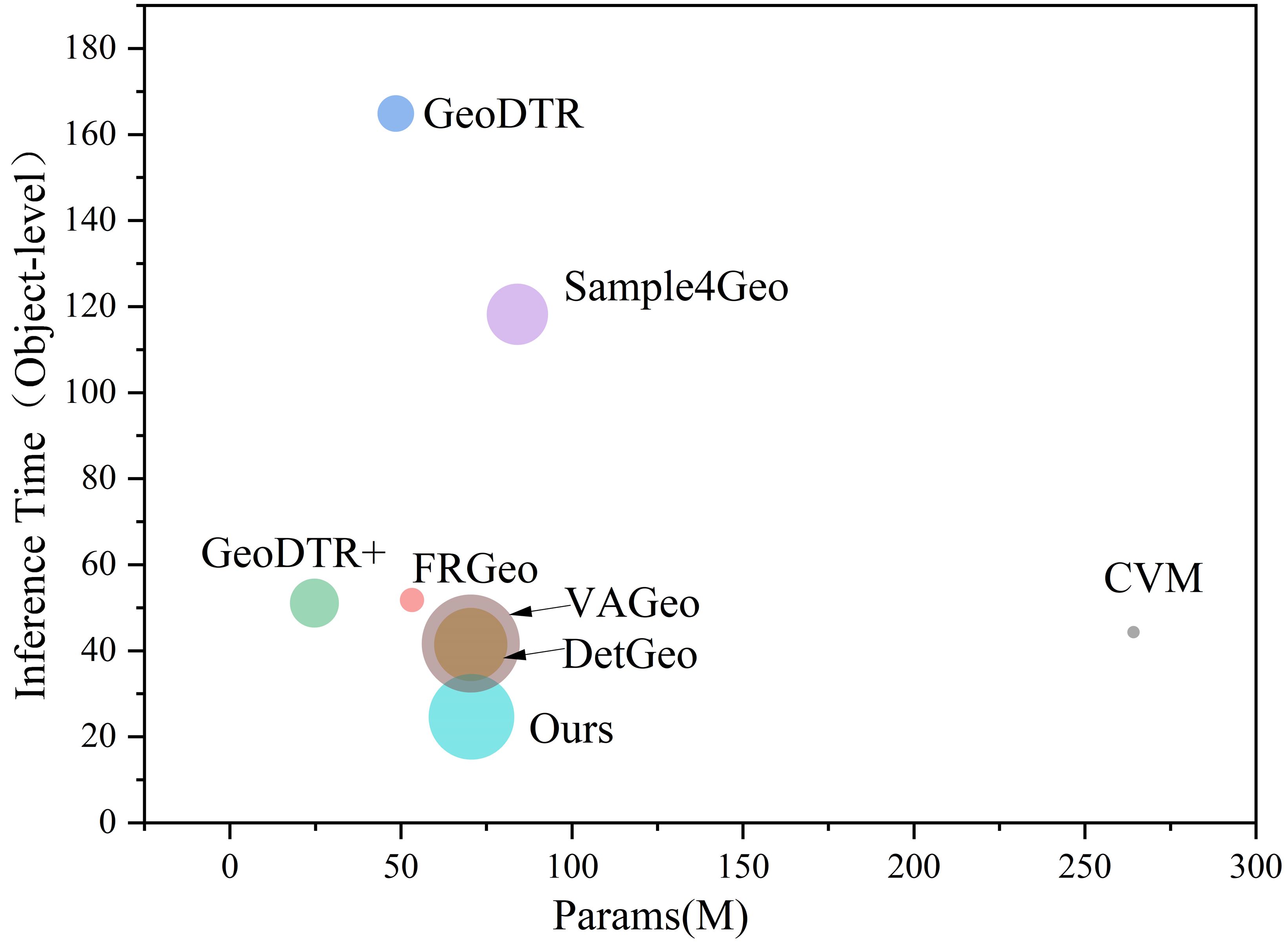}
  \caption{Relationship between model parameters and inference time, measured on the validation set of the CMLocation-V1. }
  \label{fig:Varing_query_object}
\end{figure}

\textbf{(4) Localization Speed.}  
To further investigate the trade-off between model complexity and inference efficiency, we compare the parameter counts and localization speeds of different methods in Fig.~\ref{fig:Varing_query_object}. The GeoDTR+ method has the smallest parameter count and achieves fast inference. MOGeo has a comparable number of parameters to GeoDTR+, but achieves faster inference. While VAGeo has few parameters, it is designed to localize a single object per inference, leading to increased time when handling multiple objects.

\begin{table}[htbp]
\centering
\small
\caption{Ablation performance (\text{accuracy}\%)  comparison on CVOGL-Drone and CMLocation-V2 datasets.}
\begin{tabular}{lcccc}
\toprule
\multirow{2}{*}{Method} & \multicolumn{2}{c}{Test} & \multicolumn{2}{c}{Validation} \\
\cmidrule(lr){2-3} \cmidrule(lr){4-5}
 & acc@0.25 & acc@0.5 & acc@0.25 & acc@0.5 \\
\cmidrule(lr){2-5}
& \multicolumn{4}{c}{CVOGL-Drone} \\
\cmidrule(lr){2-5}
\textbf{Ours} & 66.39 & 59.71 & 65.11 & 58.72 \\
w/o $L_s$     & 65.98 & 59.30 & 61.21 & 54.82 \\
w/o CVMF      & 65.78 & 59.92 & 62.62 & 58.40 \\
w/o MOPE       & 54.88 & 50.05 & 54.60 & 48.21 \\
\midrule
& \multicolumn{4}{c}{CMLocation-V2} \\
\cmidrule(lr){2-5}
\textbf{Ours} & 37.87 & 36.63 & 39.16 & 37.40 \\
w/o $L_s$     & 37.56 & 35.85 & 38.82 & 36.74 \\
w/o CVMF      & 37.19 & 35.57 & 37.45 & 35.73 \\
w/o MOPE       & 32.74 & 31.12 & 33.00 & 31.54 \\
\bottomrule
\end{tabular}
\label{tab:performance_CVOGL}
\end{table}

\subsection{Ablation Study}
To validate the effectiveness of our proposed components (MOPE, CVMF, \( L_s \)), we conducted a series of experiments by sequentially removing each component. We performed these experiments on two challenging datasets: CMLocation-V2 (the \emph{Ground}\textrightarrow\emph{Satellite} scenario) and CVOGL-Drone (the \emph{Drone}\textrightarrow\emph{Satellite} scenario), with the results presented in Table~\ref{tab:performance_CVOGL}.
As observed, removing any of the three components results in varying degrees of performance degradation across both datasets, which underscores the individual contribution of each component to the final performance. Notably, the removal of the MOPE module leads to the most substantial performance drop. For instance, on the CVOGL-Drone test set, the primary metric acc@0.25 drops by 11.51\%, and on CMLocation-V2 test, it decreases by 5.13\%. This significant decline, far exceeding the impact of removing other modules, strongly validates our central hypothesis and demonstrates the critical role of the proposed impulse-based positional encoding in achieving accurate multi-object localization.

 \begin{figure}
    \centering
  \includegraphics[width=1\linewidth]{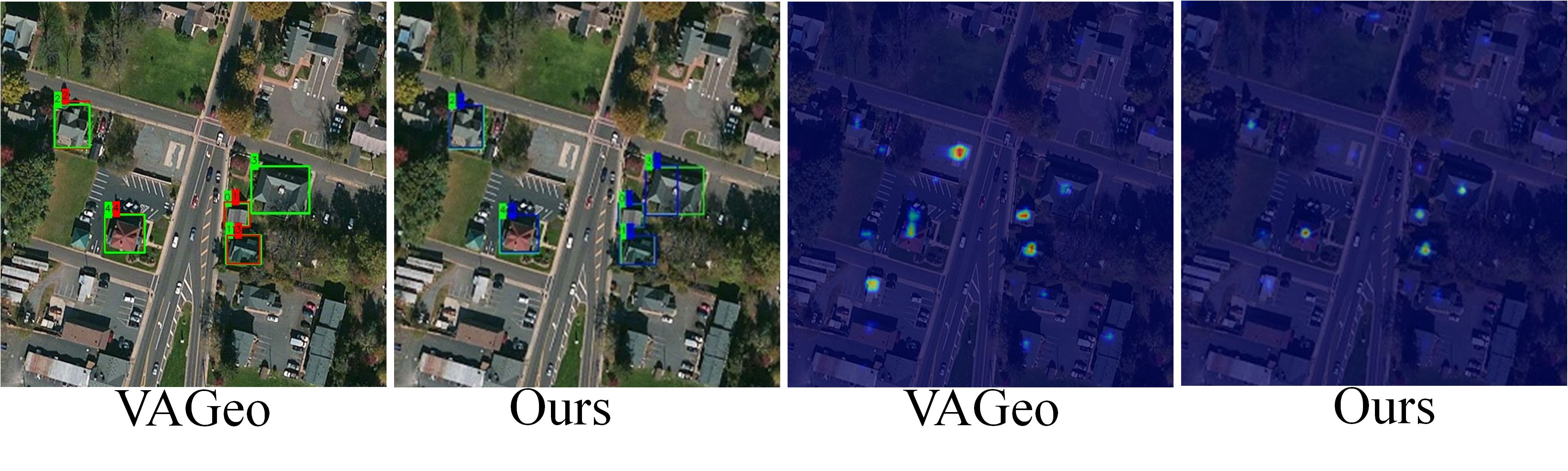}
  \caption{Detection result visualization. Green represents the GT boxes, while the red and blue bounding boxes correspond to the predictions of VAGeo and our method, MOGeo, respectively.
}
  \label{fig:detection_visual}
\end{figure}
 \subsection{ Visualization Analysis} 
 To further analyze the proposed method and compare the attention regions of different models, we present the detection results and corresponding heatmaps in Fig.~\ref{fig:detection_visual}. In the detection results visualization, numbers denote object indices, while green, red, and blue bounding boxes represent GT bounding boxes, VAGeo predictions, and MOGeo (ours) predictions, respectively. MOGeo correctly localizes all query objects, while VAGeo identifies only two (indices 2 and 4). In the heatmap visualizations, MOGeo precisely attends to objects of interest while suppressing background noise. In contrast, VAGeo exhibits spatial misalignment between its attention and objects, with attention partially shifting to non-object regions. This comparison demonstrates superior discriminative ability and robustness of MOGeo in multi-object localization.

\section{Conclusion and Limitations}
In this work, we address the limitation that existing cross-view geo-localization methods are restricted to single-object settings and fail to meet the practical demands of multi-object localization. To this end, we introduce a novel task of cross-view multi-object geo-localization. To facilitate research research in this direction, we have constructed the  CMLocation benchmark, including both V1 and V2 versions, and proposed a MOGeo multi-object localization algorithm. 
Experimental results demonstrate that the proposed method achieves consistent improvements in multi-object localization. However, limitations remain in accurately aligning query objects with their corresponding targets and localizing them precisely, particularly in complex scenarios. These findings highlight directions for future improvements.

\section{Acknowledgement}
This work was supported in part Guangdong Basic and Applied Basic Research Foundation under Grant 2026A1515011137, in part by the Key Project of Department of Education of Guangdong Province under Grant 2023ZDZX1016, and in part by Shenzhen Science and Technology Program under Grant JCYJ20240813142510014 and Grant 20220810142553001.

{
    \small
    \bibliographystyle{ieeenat_fullname}
    \bibliography{main}
}

\end{document}